# Personalization Effect on Emotion Recognition from Physiological Data: An Investigation of Performance on Different Setups and Classifiers

Varvara Kollia

**Abstract**—This paper addresses the problem of emotion recognition from physiological signals. Features are extracted and ranked based on their effect on classification accuracy. Different classifiers are compared. The inter-subject variability and the personalization effect are thoroughly investigated, through trial-based and subject-based cross-validation. Finally, a personalized model is introduced, that would allow for enhanced emotional state prediction, based on the physiological data of subjects that exhibit a certain degree of similarity, without the requirement of further feedback.

## 1 INTRODUCTION

Emotion recognition is a rapidly increasing field today, as automation and personalization turn out to be some of the key elements in most human-computer interaction (HCI) systems. The problem of machine emotional intelligence is very broad and multi-faceted; one of its challenges being the very fact that is hard to define it in an unambiguous way. There is no unique definition of emotion, and there is neither a specific method nor a particular required dataset that is guaranteed to capture it.

One of the most popular emotion definitions is the one of the six basic emotions by Paul Ekman [1]. The original six emotions he proposed are: anger, disgust, fear, happiness, sadness and surprise. Another very popular approach is the 2-dimensional emotion map, where each emotional state is projected on the orthogonal axes of valence and arousal [2]. A third dimension can be added to this space with the axis of dominance, see [3] and its related references. The axis of arousal corresponds to the variation of the level of calmness or excitement towards a stimulus. Valence is a measure of the degree of happiness or sadness the subject feels or how attracted or repulsed they are to an external factor/event. Finally, dominance represents the level of empowerment. Representative emotional states in the 3D model of valence, arousal, and dominance can be found in [3, 4]. For the most part of this work, we will be working on the 2D valence-arousal space.

## 2 METHODOLOGY OVERVIEW

### 2.1 Dataset

The DEAP dataset [5] is one of the few publicly available datasets that aims to recognize the emotional state of the subject based on their electroencephalogram (EEG) and peripheral physiological sensors. The experiment took place in a controlled setup and it consists of two parts. The first part contains the online self-assessment ratings for 120 one-min extracts of music videos, rated by 14-16 volunteers based on arousal, valence and dominance. Using as stimuli a selection of 40 of those pre-rated music video clips, the EEG and the other biosensor data of 32 volunteers were recorded, along with partial frontal face recordings and their rankings on valence, arousal, dominance, liking and familiarity were collected.

32 EEG channels were collected from each of the subjects. In addition, peripheral physiological signals were recorded. Part of the data were preprocessed (downsampled, filtered, reordered and the artifacts were removed) [6]. For this part, the blood volume pressure (BVP) from the plethysmograph, the galvanic skin response (GSR), the temperature (T), the respiration amplitude (RESP), two electromyograms ,- Zygomaticus Major EMG and Trapezius EMG-, and two electrooculograms (EOG),-horizontal and vertical-, were used. The final dataset consists of the 40-channel physiological recordings of 32 subjects for 40 videos. The self-assesment is in terms of valence, arousal, dominance and liking.

### 2.2 Problem SetUp

The typical block diagram of a classification problem is shown in Figure 1. Starting from the raw data, digital signal processing techniques [7] are usually used to remove noise and artifacts from the data.

---

• *Author is with Intel Corp. 2200 Mission College Blvd, Santa Clara, CA 95054. E-mail: Varvara.Kollia@intel.com*

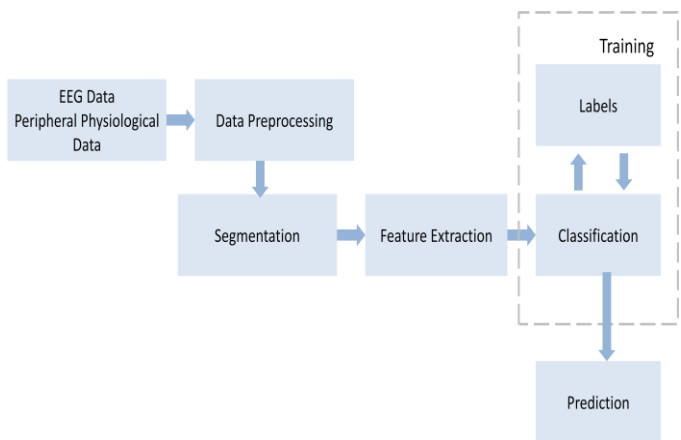

Fig.1. Block diagram of our classification problem.

Our data have been preconditioned, as explained in [6]. All data have been downsampled to 128 Hz and the 60-sec trials were reordered in terms of video-order. Artifacts were removed and the data were prefiltered, as applicable.

For the temporal segmentation process, a uniform window of 1sec was applied. There is no overlap between adjacent windows.

The feature extraction is used to reduce the size (dimensionality) of the problem and generate more representative characteristics of the problem that would lead to higher accuracy in classification. As the features are hand-designed, special attention should be given to their extraction; therefore the process we follow is described in detail in the next session. The features are used as an input to the classification algorithm. The estimated (predicted) classes are consequently compared to the actual classes to fine tune the algorithm's parameters. In the prediction phase, the learned algorithm is used to assign labels to unlabeled data. An alternative approach would be to use directly the raw data to generate abstract features (data representations) automatically with deep learning methods, as in [4,8,9].

As ground-truth here, we use the self-assessments of the subjects; i.e. the rankings they gave to each video. We mainly focus on the valence/arousal rankings for each of which a scale of 1..9 was assigned during the experiment. For the binary classification problem, the labels are mapped to High/Low valence (or arousal) depending on the subject's ranking. Specifically, if the score is less than 4.5 we consider the label to be that of low valence, otherwise it is considered high.

The 2D problem of valence and arousal is initially addressed. A typical map of emotions to the valence/arousal axis can be found in [3]. The self-assessments are mapped to a 4-label classification problem corresponding to 1 of the 4 quadrants of the valence/arousal map centered around the following four emotional states [10]: sadness, anger, pleasure, joy. The valence/arousal axes are consequently decoupled, to improve accuracy and speed of convergence. A secondary mapping on the 3D space of valence/arousal/dominance is used for comparison purposes. Representative emotional states for this space by [3,4] are shown in Fig.2. Dashed lines represent the axis of dominance.

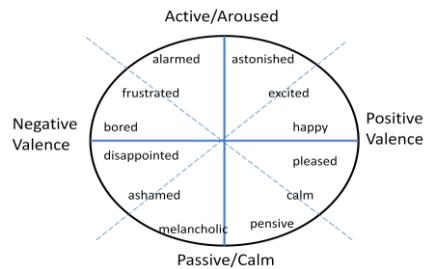

Fig.2 Emotion decomposition on its major axes.

The main classifier used is Random Forests [11]. Partial results from other classifiers [12] (namely Support Vector Machines, Decision Tree, Neural Net) are presented for comparison purposes.

### 2.3 Feature Extraction

EEG signals measure the change of the electric potentials in brain neurons, caused by changing ionic currents [13]. They are usually described by their energy or spectral power in different frequency bands. The most commonly used frequency bands are the following [14]: Delta (1-3) Hz, Theta (4-7) Hz, Alpha (8-13) Hz, Beta (14-30) Hz and Gamma (31-50) Hz. For each EEG signal, we derive the spectral power on these five frequency ranges. The power spectral density (PSD) is computed directly from the Fast Fourier Transform (fft) of the signal for each frequency band, after mapping our temporal window to the corresponding uniform frequency window. Equivalently, the Welch method/periodogram could be used [7]. Additionally, we derive the mean, the standard deviation (std), the simple square integral (SSI) and the mean frequency of the signal [15]. A total of 288 features are derived for all EEG channels.

The GSR signal measures changes in the electrodermal activity of the subject. The signal is characterized by occasional abrupt changes (startles) in response to a stimulus [16]. Its main characteristics are the rise time in the interval of the startles and the signal amplitude. For this study, the mean and the standard deviation of the rise time, along with the min, the max and the central frequency of the GSR signal are extracted for each segment.

The plethysmograph records the blood volume pressure (BVP) of the subject. From the BVP, by detecting the peaks of the signal, we can derive the inter-beat interval (R-R), the heart rate (HR) and the heart rate variability (HRV). A summary of the most popular features derived from BVP/HR measurements can be found in [17]. In particular, R-R is inversely proportional to HR. HRV is the difference between consecutive R-R intervals. Constant interpolation is used to define HRV at all times (and frequency points) of interest. An alternative would be to use (e.g. linear) interpolation and map the actual (recorded) times to the uniform grid (constant time-step) used by the Discrete Fourier Transform. Furthermore, it is straightforward to get the squared differences (SD) between successive inter-beat intervals (SSD) and the percentage of occurrences of RR variations that are greater than 50ms in a window (pNN50). Moreover, the mean and the standard deviation of the following signals are estimated: inter-beat intervals, HR, HRV, SD, SSD. Finally, the power spectral density in four frequency bands is estimated. The frequency range for each band is defined in [17] as $f \leq 0.04 Hz$ for the ultra low frequencies (ULF), $f \in (0.04, 0.15] Hz$ for the low frequency (LF) range, $f \in (0.15, 0.4] Hz$ for the high frequencies (HF) and $f > 0.4 Hz$ for the ultra high frequencies (UHF). The number of features derived from the plethysmograph for this study is 19.

The respiration belt provides the respiration signal amplitude. The mean and the standard deviation of the signal and its first derivative are calculated, along with the SSI, the min and the max value of the signal per window. In frequency domain, the spectral centroid (mean frequency) that corresponds to the breathing rhythm,[5,17] are derived.

The remaining sensor data come from the temperature (T) sensor, the EOG and the EMG sensors. Only time-domain features (mean, std and SSI) are derived for the T sensor. EOG is used to track eye movement [18]. Two EOG sensors are used; horizontal EOG and vertical EOG. EMG [17] is used to record muscle activity as a response to neuron stimulation. Two EMG channels are available, zygomaticus major and trapezius EMG. Mean, std, SSI, and peak frequency (excluding the DC component) are used as features for the two EOG and the two EMG channels. The mean and the std of the first derivative of the EOG signal are also derived, resulting in 20 features for the eye-movement tracking components of the experiment.

A total of 348 handcrafted features in time- and frequency-domain are extracted. Non overlapping windows of 1 sec are used to define the features and the labels. The labels are extracted directly from the self-assessments of the volunteers.

Prototype code was used for the feature extraction in Matlab [19] using the Signal Processing Toolbox [20] for peak extraction and fft, as well as in R [21] for the classification part.

## 3 RESULTS

### 3.1 Multinomial Classification with Random Forest

#### 3.1.1. Subject-Independent Results

In this section we will be addressing the problem of the classification with Random Forests (RF) [11], one of the very popular classifiers in machine learning, for the 4-label classification setup. The algorithmic implementation used can be found in [22]. Please note that the 4-levels in the valence/arousal space correspond to its four quadrants. The mapping was based using as centre the middle values of the ranking intervals for valence and arousal, i.e. (4.5, 4.5). Thus the 4 classes correspond to the following values for the pair (valence, arousal): (L L),(L H),(H L),(H H), where H stands for High and L stands for Low. The 8 classes equivalent for the 3D space of (valence/arousal/dominance) can be formed in an analogous manner.

The Random Forest classifier is an ensemble classifier, that trains a large number of decision trees and assigns labels with a majority voting scheme. The Out-Of-Bag error (OOB) is a metric of the accuracy of the classification on the training set. In particular, for each tree, a part (30%) of the data is left out and used for cross-validation purposes. The mean classification error generated in this manner is the OOB error.

Table I: Confusion Matrix for 4-level Classification

| Qb \ Est | Class 1 | Class 2 | Class 3 | Class 4 | OOB |
|---|---|---|---|---|---|
| Class 1 | 31823 | 689 | 845 | 222 | 0.05 |
| Class 2 | 4221 | 12068 | 799 | 237 | 0.30 |
| Class 3 | 3170 | 580 | 13882 | 323 | 0.23 |
| Class 4 | 2802 | 617 | 539 | 7823 | 0.34 |
| OOB Error 18.70% | | | | | |

The data are organized in a 80640x349 matrix, where 80640 is the number of training samples (40x32x63) and 349 is the number of features (348) plus the label. The data are shuffled to reduce any ordering effect. The unscaled data suffer from subject/sensor bias, therefore the results are not reliable.

To remove the subject bias and generate more comparable features that would give more reliable results, we scale the data as follows. The data for each subject are normalized per sensor for all videos. In this manner, we preserve the sensor variability but remove the bias for each subject and the very different value ranges of each sensor. This is shown schematically below in Figure 3, where the arrow below each column point to the data that will be normalized.

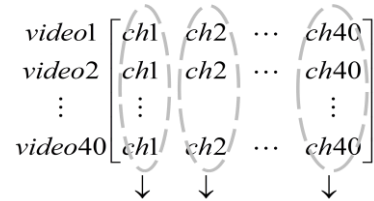

Fig.3 Illustration of data normalization for one subject.

The number of attributes we use for each random forests call is 19, approximately equal to the square root of the total number of features (348), which has been reported to generate the most accurate results in terms of generalization error [12]. The results are reported for a forest of 100 trees, since the algorithm has (almost) converged at that point to the solution. Slow small improvement is observed with the addition of more trees.

Table I shows the confusion matrix of the classification with the rows corresponding to the actual (observed) labels and the columns to the estimated ones. Random Forests is biased towards the majority class (Low Valence, Low Arousal), therefore the OOB, which is the mean classification error of the unused samples for each tree, is significantly smaller than the error for the remaining classes. Very similar OOB error is observed for the 8-level classification, as shown in Table II, for the equivalent problem in the 3D dimensional space of valence/arousal/dominance.

Table II: Confusion Matrix for 8-level Classification

| Ob\Est | Class 1 | Class 2 | Class 3 | Class 4 | Class 5 | Class 6 | Class 7 | Class 8 | Err |
|---|---|---|---|---|---|---|---|---|---|
| Class 1 | 29140 | 15 | 310 | 45 | 81 | 269 | 12 | 179 | 0.03 |
| Class 2 | 1219 | 1995 | 81 | 36 | 44 | 126 | 1 | 26 | 0.44 |
| Class 3 | 2644 | 4 | 9527 | 66 | 150 | 230 | 12 | 93 | 0.25 |
| Class 4 | 1201 | 38 | 235 | 2921 | 24 | 66 | 0 | 114 | 0.37 |
| Class 5 | 1629 | 4 | 124 | 41 | 5580 | 319 | 12 | 103 | 0.29 |
| Class 6 | 1543 | 5 | 230 | 8 | 84 | 8104 | 2 | 167 | 0.20 |
| Class 7 | 945 | 14 | 176 | 40 | 39 | 40 | 2015 | 70 | 0.40 |
| Class 8 | 1655 | 18 | 182 | 53 | 57 | 183 | 5 | 6289 | 0.26 |
| OOB Error 18.77% | | | | | | | | | |

The most important features based on Gini index [12,22] for the 4-labels setup are the following (from most important to least important one): Standard deviation of the first derivative of the EOG channels and of the RESP signal, the power spectrum for EEG channels 1,11 and 15 in the gamma frequency band, the mean, the power spectrum in the ULF band for the HR and the standard deviation of the RR interval (HR), the power spectrum for EEG channels 31 and 1 in the beta frequency band, the standard deviation of the HRV signal and of the second EMG channel and, finally, the power spectrum in ULF and LF bands of the HRV. We get similar results for the 8 labels setup, where we see more prominently the effect of the Respiration Signals and GSR characteristics (mean, min/max).

### 3.1.2. Personalized (Subject-Dependent) Models

The problem of inter-subject variability is well known in emotion recognition, [5,13,23], thus for the remaining of this work we develop personalized models for each subject. Figure 4 can be used to interpret the class enumerations of the results' Tables of this section. More specifically, Table III shows the OOB error/subject. Significant error reduction is observed. However, the results show clearly greater accuracy towards the majority class, as expected. The N/A indication in the results means that there were no mappings from the subject's rankings to the class in question.

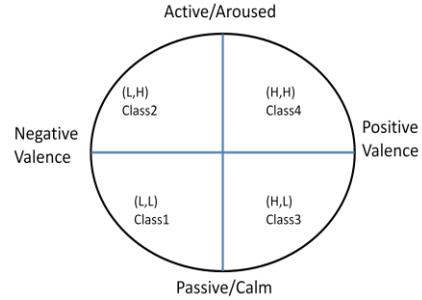

Fig. 4 Mapping of classes in 2D valence/arousal space.

To overcome the problem of class imbalance, we employee the stratified sampling technique, which overcomes this problem by forcing the classifier to choose more balanced samples during the training phase. This is a balanced RF technique [24]. Here, we force the training set to consist of samples where the effect of the most populous class is undermined; i.e., the majority class is undersampled so that the minority class is represented substantially in the bootstrapped samples; here we chose this representation to be at least a fraction of it (e.g. 1/6).

Tables IV and V summarize the results. In particular, Table IV shows the comparative results of the personalized Random Forests classifier for the first five subjects with and without stratified sampling. Every second line (denoted by S in italic) contains the balanced RFs result. The improvement accuracy is evident, both for the overall OOB, as well as for the minority classes. Similar improvement in results is observed for all subjects. All the results in Tables III-V, as well as in the next sections, are in (%), unless denoted otherwise.

Table III: Error (%) for Personalized Model/Subject with RFs

| Subject | OOB (%) | Class 1 | Class 2 | Class 3 | Class 4 |
|---|---|---|---|---|---|
| Subject 1 | 9.76 | 2.38 | 28.31 | 8.66 | 10.23 |
| Subject 2 | 16.03 | 2.94 | 32.98 | 18.59 | 38.89 |
| Subject 3 | 14.52 | 41.53 | 0.46 | 35.98 | 30.39 |
| Subject 4 | 17.66 | 14.95 | 42.86 | 37.04 | 5.32 |
| Subject 5 | 19.21 | 1.27 | 44.25 | 30.48 | 33.79 |
| Subject 6 | 15.04 | 5.38 | 9.42 | 33.33 | 65.87 |
| Subject 7 | 9.17 | 1.35 | 9.52 | 31.43 | 17.46 |
| Subject 8 | 10.75 | 1.21 | 31.11 | 15.71 | 23.02 |
| Subject 9 | 8.17 | 0.56 | 17.99 | 14.11 | 16.87 |
| Subject 10 | 10.95 | 6.11 | 9.52 | 11.90 | 22.49 |
| Subject 11 | 5.99 | 5.95 | 2.12 | 8.47 | 12.24 |
| Subject 12 | 14.05 | 5.32 | 46.03 | 11.32 | 73.81 |
| Subject 13 | 8.37 | 10.52 | 25.00 | 2.01 | 28.57 |
| Subject 14 | 11.07 | 4.63 | 18.86 | 4.02 | 36.83 |
| Subject 15 | 6.83 | 8.10 | 6.22 | 3.84 | 11.90 |
| Subject 16 | 6.27 | 4.06 | 4.17 | 11.98 | 4.10 |
| Subject 17 | 12.02 | 1.32 | 25.20 | 19.84 | 16.14 |
| Subject 18 | 11.19 | 0.68 | 27.34 | 21.16 | 15.08 |
| Subject 19 | 15.87 | 1.19 | 50.00 | 22.22 | 22.22 |
| Subject 20 | 14.96 | 2.83 | 43.49 | 23.28 | 58.73 |
| Subject 21 | 11.03 | 1.83 | 17.14 | 14.15 | 49.74 |
| Subject 22 | 13.73 | 4.87 | 27.66 | 8.16 | 42.06 |
| Subject 23 | 4.56 | 3.37 | 4.46 | 7.14 | N/A |
| Subject 24 | 10.56 | 11.43 | 38.10 | 2.22 | 53.97 |
| Subject 25 | 13.61 | 3.36 | 34.47 | 11.32 | 43.65 |
| Subject 26 | 13.53 | 2.17 | 30.16 | 87.30 | 14.29 |
| Subject 27 | 11.63 | 0.19 | 26.19 | 39.68 | 32.94 |
| Subject 28 | 7.62 | 2.68 | 11.29 | 19.44 | 7.50 |
| Subject 29 | 17.34 | 4.37 | 52.38 | 30.84 | 23.33 |
| Subject 30 | 13.37 | 2.69 | 44.76 | 50.00 | 16.35 |
| Subject 31 | 9.64 | 17.30 | 3.84 | 12.70 | N/A |
| Subject 32 | 6.87 | 5.42 | 9.37 | 5.14 | 28.57 |
| Mean (OOB) | 11.60 | | | | |
| Std (OOB) | 3.55 | | | | |

Table IV: Sample of RFs vs Balanced RFs Results for % Error

|  | OOB (%) | Class 1 | Class 2 | Class 3 | Class 4 |
|---|---|---|---|---|---|
| Subject 1 | 9.76 | 2.38 | 28.31 | 8.66 | 10.23 |
| *Subject 1 S* | *7.58* | *1.70* | *24.07* | *5.48* | *8.29* |
| Subject 2 | 16.03 | 2.94 | 32.98 | 18.59 | 38.89 |
| *Subject 2 S* | *12.34* | *2.38* | *25.40* | *12.93* | *31.75* |
| Subject 3 | 14.52 | 41.53 | 0.46 | 35.98 | 30.39 |
| *Subject 3 S* | *9.05* | *27.51* | *0.46* | *19.58* | *18.14* |
| Subject 4 | 17.66 | 14.95 | 42.86 | 37.04 | 5.32 |
| *Subject 4 S* | *12.90* | *13.76* | *29.52* | *21.96* | *4.20* |
| Subject 5 | 19.21 | 1.27 | 44.25 | 30.48 | 33.79 |
| *Subject 5 S* | *11.63* | *0.71* | *26.39* | *18.41* | *21.09* |

Table V shows the comparative results with and without stratified sampling for the 4-level and the 8-level setups described above. The first two columns of this table correspond to the OOB/subject for the 2D and the 3D mapping, respectively, whereas the last two columns are the corresponding results for the balanced RFs. The 3D mapping results (the 3rd axis is the axis of dominance) are shown in italic. The 2D accuracy on the training set for the personalized models is comparable for both the 2D and the 3D settings, with the 3D setup being a little worse. Balanced RFs give more accurate results in all cases, with an improvement of 3.86% for the mean OOB and 0.88% for the std for the 4-label problem (2D). Respectively, the improvement is larger for the 8-Labels setup (3D), with 5.55% smaller OOB and 1.53% smaller standard deviation.

Table V: OOB (%) Comparison for RFs vs balanced RFs

| Subject | OOB (4 L) | *OOB (8 L)* | OOB (4 L) Strat | *OOB (8 L) Strat* |
|---|---|---|---|---|
| Subject 1 | 9.76 | *10.44* | 7.58 | *4.05* |
| Subject 2 | 16.03 | *15.79* | 12.34 | *6.94* |
| Subject 3 | 14.52 | *13.57* | 9.05 | *8.06* |
| Subject 4 | 17.66 | *22.90* | 12.90 | *15.16* |
| Subject 5 | 19.21 | *13.10* | 11.03 | *7.46* |
| Subject 6 | 15.04 | *15.99* | 11.03 | *8.10* |
| Subject 7 | 9.17 | *8.57* | 6.23 | *4.44* |
| Subject 8 | 10.75 | *13.45* | 6.98 | *6.27* |
| Subject 9 | 8.17 | *8.97* | 4.72 | *3.89* |
| Subject 10 | 10.95 | *9.09* | 7.02 | *6.27* |
| Subject 11 | 5.99 | *7.22* | 2.70 | *3.25* |
| Subject 12 | 14.05 | *17.50* | 8.97 | *8.81* |
| Subject 13 | 8.37 | *10.44* | 6.27 | *5.67* |
| Subject 14 | 11.07 | *13.02* | 7.14 | *7.50* |
| Subject 15 | 6.83 | *5.75* | 4.44 | *4.96* |
| Subject 16 | 6.27 | *6.47* | 4.29 | *4.13* |
| Subject 17 | 12.02 | *10.08* | 5.87 | *6.55* |
| Subject 18 | 11.19 | *11.71* | 6.94 | *4.76* |
| Subject 19 | 15.87 | *17.86* | 10.95 | *6.55* |
| Subject 20 | 14.96 | *16.75* | 10.16 | *10.40* |
| Subject 21 | 11.03 | *15.04* | 8.33 | *6.63* |
| Subject 22 | 13.73 | *11.98* | 9.09 | *6.59* |
| Subject 23 | 4.56 | *4.80* | 3.06 | *3.61* |
| Subject 24 | 10.56 | *14.29* | 5.63 | *7.26* |
| Subject 25 | 13.61 | *13.13* | 8.65 | *8.37* |
| Subject 26 | 13.53 | *15.20* | 8.93 | *9.29* |
| Subject 27 | 11.63 | *10.95* | 7.06 | *6.55* |
| Subject 28 | 7.62 | *8.13* | 4.96 | *4.72* |
| Subject 29 | 17.34 | *15.32* | 12.18 | *8.73* |
| Subject 30 | 13.37 | *14.48* | 8.37 | *9.13* |
| Subject 31 | 9.64 | *11.19* | 7.18 | *5.36* |
| Subject 32 | 6.87 | *10.32* | 10.24 | *6.03* |
| Mean | 11.60 | *12.28* | 7.74 | *6.73* |
| Std | 3.55 | *3.92* | 2.67 | *2.39* |

### 3.1.3 Comparison to Other Classifiers

Compared to Random Forests a decision tree [25] gives much worse accuracy, with a mean 10-fold cross validation error [26], on the 4-labels training set, equal to 58.36%. This is expected, since RFs are a more general and modular classifier compared to decision trees. The decision tree was trained on the 10 dominant features based on the Random Forests feature ranking on Gini Index.

Choosing SVMs as classifier [27] with radial kernel, 10-fold cross validation on the training set yields 37.43% error, which is higher compared to the OOB error on the training set. The SVM classifier outperforms the decision tree classifier by ~21% in accuracy. The SVM parameters were optimized beforehand, and these optimal parameters were employed in the classification (cost=1000, gamma =0.1).

For a slightly modified setup, corresponding to the binary classification problem of High/Low valence, using a shallow neural network [28] with 1 hidden layer with 5 neurons, for the personalized models, 10-fold cross validation results give a mean error ~20% on the training set. For convergence reasons, the first 9 dominant features were used as inputs and the network topology was defined in terms of the number of input/output nodes. The activation function was the logistic function and the network parameters (weights, biases) were trained with the modified globally convergent back propagation algorithm. These results are worse than the corresponding Random Forests results, as we will see in the next section.

### 3.2 Binary Classification with Random Forest – Personalized Models

Inter-subject variability is a bottleneck in emotion classification from physiological data. Leave-one (subject)-out cross validation on 4 Labels gives very low accuracy, with (in %) mean cross-validation error 60.76, and standard deviation 9.73. The results are comparable to the ones obtained in [4], taking into consideration that [4] employs a 3-Label setup. However, if we use as similarity criterion the ranking preferences of the subjects and take into account only the personalized models built on the most similar participants, then we can get the same (or better) accuracy for most subjects. Specifically, if we use only the subjects for which the Pearson Correlation coefficient [7] is ρ>0.35, then the mean cross validation error decreases to 58.87%. In the remaining part of the section, we will look more into this effect.

Decoupling the problem into the axis of valence and arousal, we end up with two binary classification problems that combined can reproduce the 4-label problem described above. A random forest is trained for each subject and tested on leave-one (trial) –out scenario. Thus, for each subject the model is trained on the 39 videos based on the subject labeling, and tested on the remaining one. The process is repeated until all videos are tested. The statistics of this process for all subjects in (%) can be seen in Table VI. The large value of the mean of the standard deviation is due to outliers towards the higher error values. Since for every subject we get a 63x40 error matrix, the statistics for the mean for all subjects (i.e. a matrix of dimensions 63x40x32), involve the {mean, std, median} value of the {mean, std} for all subjects.

Table VI: Statistics of Error (%) from Leave-One (Trial) -Out Cross Validation

(a)

| Valence | | |
|---|---|---|
| | Mean | Std. Dev. |
| Mean | 20.70 | 27.10 |
| Median | 18.79 | 24.60 |
| Std. Dev. | 7.92 | 9.89 |

(b)

| Arousal | | |
|---|---|---|
| | Mean | Std. Dev. |
| Mean | 22.58 | 30.11 |
| Median | 19.86 | 28.15 |
| Std. Dev | 10.76 | 10.32 |

Finally, Table VII illustrates the statistics of the average error (%), that would correspond to the actual real-life scenario, where we truncate for each video, all the classified values we get from the corresponding windows to end up with one ranking. Then, this value is compared to the self-assessment value for the video from the subject. The process is repeated for all videos for all subjects. The results are consistent (or better) with results reported in literature, see eg [5]. The mean accuracy for both valence and arousal is ~65%, whereas the results for valence have a slightly smaller spread.

Table VII: Statistics of Truncated Error (%) from Leave-One (Trial)-Out Cross Validation

| | Valence | Arousal |
|---|---|---|
| Mean Error | 35.16 | 35.15 |
| Median Error | 33.75 | 36.25 |
| Std. Dev. | 10.28 | 13.84 |
| Mean Accuracy | 64.84 | 64.85 |

In Figure 5 the histograms of the accuracy for the different subjects is illustrated for the personalized leave-one (video)-out cross validation scheme.

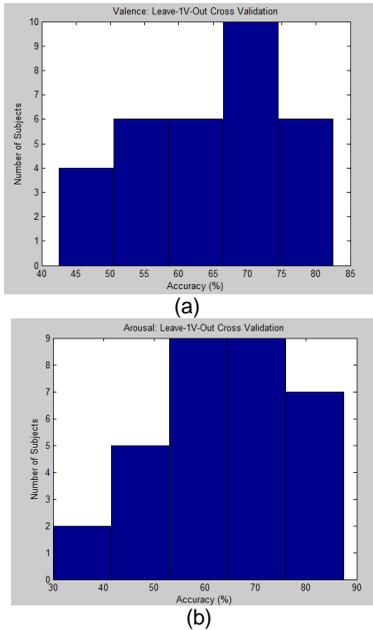

Figure 5. Histograms of Accuracy for Leave-One (Trial)-Out Cross Validation for (a) Valence and (b) Arousal.

Next, a clustering method [12] with similarity metric the Pearson correlation coefficient is used, so that every cluster consists of at least two subjects. In order to achieve this, we choose $\rho>0.35$ for the L/H valence-problem and $\rho>0.3$ for the corresponding one for arousal. For each subject, in the leave-one (subject) out cross-validation scheme, the models only from the subjects belonging to the same cluster (nearest neighbors, kNN algorithm [12]) will be used, which reduces significantly the computational time. For the prediction, a majority vote is taken among the subject's neighbors. It is noteworthy that we get similar results after training the RF model on the entire dataset comprised of the inter-cluster data. The results are summarized in Table VIII and compare favorably to the previously reported results for the subject-independent data. This means that if we can assign a subject into a group of similar subjects (people watching music clips in this case), then we can predict the subject's preferences, based on the other subjects' physiological signals without additional input from them (rankings for new clips).

Table VIII: Statistics of Error (%) from Leave-One (Subject)-Out Cross Validation

|  | Valence | Arousal |
|---|---|---|
| Mean Error | 37.37 | 43.81 |
| Median Error | 37.17 | 49.92 |
| Std. Dev. | 10.37 | 16.94 |
| Mean Accuracy | 62.63 | 56.19 |
| Min | 18.06 | 12.30 |
| Max | 57.22 | 86.74 |

The results are shown analytically in Figure 6 and Figure 7. The accuracy/subject for the binary classification problem of valence and arousal is shown in Figure 6 and the histograms of accuracy in 10-point bins (intervals) that represent the number of subjects that their accuracy lies in a certain interval are shown in Figure 7. Different shapes correspond to different intervals of accuracy (Acc); star: Acc<50%, diamond: Acc in [50,60) %, square: Acc in [60,70]%, circle: Acc > 70%.

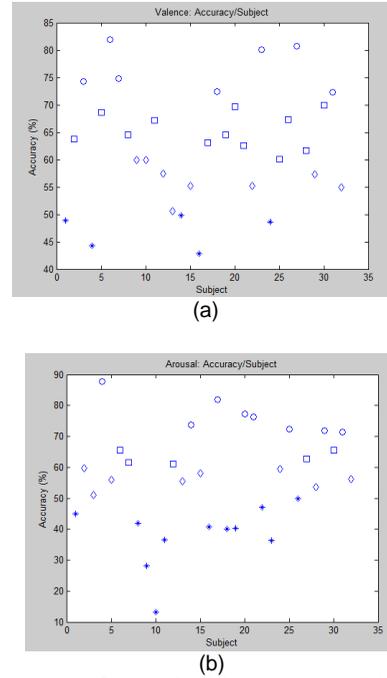

Figure 6. Accuracy/Subject for (a) Valence and (b) Arousal for prediction based on a nearest-neighbors model.

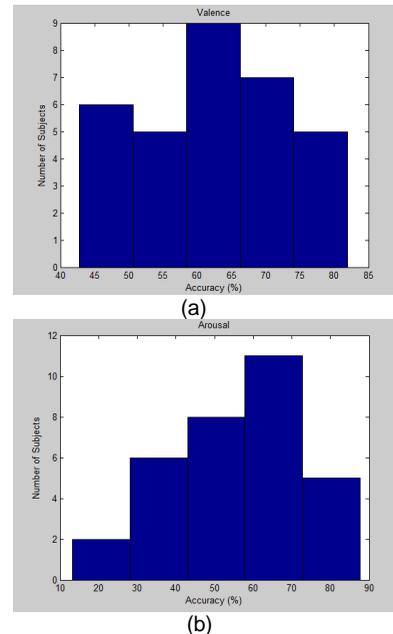

Figure 7. Histograms of prediction accuracy for (a) Valence and (b) Arousal for the problem of Figure 6.

## 4 SUMMARY

We focused our analysis on the EEG and peripheral physiological signals of the DEAP dataset. Features on time- and frequency-domain were extracted and the accuracy for different setups was examined. The features were ranked using the Gini index criterion. The greatest accuracy was achieved with personalized models on the decoupled binary valence/arousal problem. Different classifiers were briefly compared and random forests were found to be most efficient among them. The personalized setup was further investigated in terms of the efficiency to generalize based on data (subject ranking) similarity. Both scenarios of leave-one-out cross validation schemes for videos and subjects are investigated, as well as the inter-subject variability. Personalized models outperform general models significantly in terms of prediction accuracy. One notable result is that it is possible to get similar (or better) prediction accuracy using a few similar subjects compared to the personalized models built from the general population.